# The Significance of Machine Learning in Clinical Disease Diagnosis: A Review

S.M. Atikur Rahman
Department of Industrial, Manufacturing and Systems Engineering
University of Texas at El Paso
El Paso, TX 79968, USA

Sifat Ibtisum
Department of Computer Science
Missouri University of Science and Technology, Rolla, Missouri

Ehsan Bazgir
Department of Electrical Engineering
San Francisco Bay University
Fremont, CA 94539, USA

Tumpa Barai
Department of CSE
European University Bangladesh
Dhaka, Bangladesh

## ABSTRACT
The global need for effective disease diagnosis remains substantial, given the complexities of various disease mechanisms and diverse patient symptoms. To tackle these challenges, researchers, physicians, and patients are turning to machine learning (ML), an artificial intelligence (AI) discipline, to develop solutions. By leveraging sophisticated ML and AI methods, healthcare stakeholders gain enhanced diagnostic and treatment capabilities. However, there is a scarcity of research focused on ML algorithms for enhancing the accuracy and computational efficiency. This research investigates the capacity of machine learning algorithms to improve the transmission of heart rate data in time series healthcare metrics, concentrating particularly on optimizing accuracy and efficiency. By exploring various ML algorithms used in healthcare applications, the review presents the latest trends and approaches in ML-based disease diagnosis (MLBDD). The factors under consideration include the algorithm utilized, the types of diseases targeted, the data types employed, the applications, and the evaluation metrics. This review aims to shed light on the prospects of ML in healthcare, particularly in disease diagnosis. By analyzing the current literature, the study provides insights into state-of-the-art methodologies and their performance metrics.

## 1. INTRODUCTION
In the medical field, artificial intelligence (AI) plays a crucial role in developing algorithms and techniques to aid in disease diagnosis. Medical diagnosis entails determining the illness or conditions that account for an individual's symptoms and indicators, usually relying on their medical background and physical assessment. However, this process can be challenging as many symptoms are ambiguous and require expertise from trained health professionals. This becomes particularly problematic in countries like Bangladesh and India, where there is a scarcity of healthcare professionals, making it difficult to provide proper diagnostic procedures for a large population of patients. Additionally, medical tests required for diagnosis can be expensive and unaffordable for low-income individuals [1-3].

Due to human error, over diagnosis can occur, leading to unnecessary treatment and negatively impacting both the patient's health and the economy. Reports suggest that a significant number of people experience at least one diagnostic mistake during their lifetime. Several factors contribute to misdiagnosis, including the lack of noticeable symptoms, the presence of rare diseases, and diseases being mistakenly omitted from consideration [4, 5]. ML has found widespread applications in various fields, from cutting-edge technology to healthcare, including disease diagnosis. Its popularity is growing, and it is becoming increasingly utilized in healthcare to improve diagnostic accuracy and safety.

ML serves as a robust mechanism enabling machines to learn autonomously, eliminating the requirement for explicit programming. It harnesses sophisticated algorithms and statistical methods to analyze data and formulate predictions, departing from traditional rule-based systems. The accuracy of machine learning predictions heavily depends on the quality and relevance of the dataset used. Its applications span various industries, including finance, retail, and healthcare [6, 7], where it presents significant opportunities for disease diagnosis and treatment.

One of the notable features of machine learning is its continuous improvement in data prediction and classification. As more data is gathered, the prediction models become more adept at making accurate decisions. In the healthcare sector, patient datasets stored in electronic healthcare records can be leveraged to extract relevant information using ML techniques [8, 9, 10]. These algorithms aid in disease diagnosis by analyzing data and predicting the underlying causes of illnesses based on disease-causing variables extracted from electronic health records [11]. Compared to traditional bio statistical approaches, machine learning has gained popularity for tasks like classification, prediction, and clustering involving complex healthcare data. It has demonstrated exceptional results in various medical tasks, such as identifying body organs from medical images [12], classifying interstitial lung diseases [13], reconstructing medical images [14, 15], and segmenting brain tumors [15]. Overall, the use of ML in healthcare has shown great promise in advancing disease analysis, diagnosis, and treatment, showcasing its potential to transform the field by leveraging vast amounts of data for accurate and efficient healthcare solutions [16-21].

## 2. AI IN HEALTHCARE AND MEDICINE

The utilization of AI and the related technologies is becoming more widespread in both the business sector and society. This trend is now extending to the healthcare domain. The aforementioned technologies has the capacity to revolutionize various facets of patient care, as well as administrative procedures within provider, payer, and pharmaceutical entities. Machine learning is a statistical methodology utilized to effectively fit models to data and acquire knowledge through the process of training models with data. Machine learning is widely recognized as a prevalent manifestation of artificial intelligence. In the field of healthcare, precision medicine is a widely employed application of traditional machine learning. It involves the prediction of treatment outcomes for patients by considering a range of patient features and the contextual factors around the therapy.

Supervised learning is a fundamental requirement for the bulk of machine learning and precision medicine applications, wherein a training dataset is necessary to possess known outcome variables, such as the beginning of disease.

The neural network, a sophisticated variant of machine learning, has been a prominent technology in healthcare research for several decades. Its origins can be traced back to the 1960s. Neural networks have been effectively employed in categorization tasks, such as predicting the likelihood of a patient developing a specific disease. The framework adopts a perspective that analyses problems by considering the inputs, outputs, and weights of variables, sometimes referred to as "features," which establish the associations between inputs and outcomes. The most intricate manifestations of machine learning encompass deep learning, which pertains to neural network models characterized by numerous tiers of features or variables that facilitate the prediction of events. The quicker processing capabilities of contemporary graphics processing units and cloud infrastructures have enabled the discovery of several latent features inside these models, perhaps amounting to thousands. One prevalent utilization of deep learning in the healthcare field involves the identification and classification of possibly malignant lesions in radiographic images. The utilization of deep learning techniques is becoming more prevalent in the field of radiomics, which involves the identification of diagnostically significant characteristics in imaging data that beyond the capabilities of human visual perception [22]. Radiomics and deep learning are frequently utilized in the domain of oncology-focused image analysis. The amalgamation of these technologies exhibits potential for enhanced diagnostic precision compared to the preceding iteration of automated image analysis tools, commonly referred to as computer-aided detection (CAD) [23, 24].

Over the years, intelligent healthcare systems have commonly relied on centralized artificial intelligence (AI) capabilities situated in either the cloud or the data center to facilitate the learning and analysis of health data. The current centered solution in modern healthcare networks is inefficient in terms of communication latency and lacks high network scalability due to the growing volumes of health data and the proliferation of IoMT driven devices. Moreover, the dependence on a centralized server or third-party entity for data learning gives rise to significant privacy concerns, such as the potential leakage of user information and the risk of data breaches. This assertion holds special validity within the realm of e-healthcare, as health-related data is characterized by its high sensitivity and privacy, hence necessitating adherence to health standards. Furthermore, it is anticipated that in forthcoming healthcare systems, a centralized AI architecture may become less appropriate due to the decentralized nature of health data, which is dispersed across a vast IoMT network. Hence, it is imperative to adopt distributed artificial intelligence (AI) methodologies in order to facilitate the development of scalable and privacy-conscious intelligent healthcare applications at the network edge. In the present scenario, federated learning (FL) has emerged as a viable method for achieving cost-effective smart healthcare applications while enhancing privacy protection [25, 26]. From a conceptual standpoint, FL is an AI methodology that facilitates the development of AI models of superior quality. This is achieved by combining and averaging local updates obtained from numerous health data clients, such as Internet of Medical Things (IoMT) devices [27, 28]. Notably, Florida accomplishes this without necessitating direct access to the individual data stored locally. This measure has the potential to hinder the disclosure of sensitive user information and user preferences, thereby reducing the dangers associated with privacy leakage. In addition, the utilization of FL in the healthcare domain allows for the aggregation of substantial computational and dataset resources from various health data clients, thereby enhancing the quality of AI model training, particularly in terms of accuracy. This improvement may not be attainable through the implementation of centralized AI approaches that rely on smaller datasets and have limited computational capabilities [29, 30].

## 3. ML FOR DIFFERENT DISEASE DIAGNOSIS

In recent years, the proliferation of accessible hardware and cloud computing resources has ushered in a significant increase in the application of Machine Learning (ML) across various facets of human life. This span encompasses domains as diverse as leveraging ML for personalized social media recommendations to its adoption for streamlining industrial processes through automation. Among these evolving domains, the healthcare sector stands out as an industry progressively adapting to the potential of ML. The implementation of ML algorithms within healthcare holds tremendous promise due to the substantial data volume amassed for each individual patient. This reservoir of data empowers ML algorithms to proactively chart comprehensive treatment plans for patients, contributing to cost reduction and an enhanced overall patient experience. This phenomenon is particularly advantageous, positioning ML as a latent advantage within the healthcare industry. The sector grapples with an abundance of unstructured data, including patient records, historical treatment methods, and familial medical histories. By analyzing these data repositories, ML algorithms bolster healthcare professionals in predicting forthcoming health issues, thus effectively capitalizing on patients' historical data.

The rapid progression of ML technology has catalyzed the paradigm shift towards information-centric healthcare administration and delivery. Contemporary healthcare enhancement strategies, characterized by a multidisciplinary approach, in conjunction with refined imaging and genetics-informed personalized therapeutic models, hinge on the underpinning of ML-powered information systems. As such, Machine Learning is substantiating its role as an indispensable asset poised to drive significant advancements within the healthcare domain.

Various ML approaches have gained significant attention from both academics and practitioners in disease diagnosis. This section provides an overview of focusing on the application of ML models in diagnosing various types of diseases. Notably, the global relevance of COVID-19 has led to numerous studies focusing on its detection using ML since 2020, which also received priority in our investigation. We briefly discuss severe diseases like heart disease, kidney disease, breast cancer, and Dementia.

## 3.1 Dementia Classification

Alzheimer's Disease (AD) constitutes the most prevalent form of dementia necessitating extensive medical attention. AD is a chronic brain disorder with neurobiological origins that gradually leads to the demise of brain cells. This progression results in impairments to memory and cognitive abilities, eventually leading to an inability to perform basic tasks. Dementia linked to Alzheimer's manifests in various stages:

(a) Mild Cognitive Impairment: Often marked by memory lapses as individuals age, it can also evolve into dementia for some.

(b) Mild Dementia: Individuals with mild dementia experience cognitive difficulties that impact their daily routines. Symptoms include memory loss, confusion, personality changes, disorientation, and struggles with routine tasks.

(c) Moderate Dementia: This stage involves increased complexity in daily life, requiring additional care and support. Symptoms mirror mild dementia but are more pronounced. Patients may exhibit personality shifts, paranoia, and sleep disturbances.

(d) Severe Dementia: Symptoms worsen in this phase, with communication impairments and a need for constant care. Basic functions like bladder control and maintaining head position become challenging. Even simple actions, such as sitting in a chair, become unmanageable.

**Table 1: ML in Dementia Diagnosis**

| Ref. | Dataset | Model | Accuracy (%) | Specificity (%) | Recall (%) |
|---|---|---|---|---|---|
| [31] | OASIS (373 samples, 10 variables) | XGB | 85.61 | 81.40 | 77.20 |
| [32] | 169 Samples, 14 variables | RF | 92 | 88 | 88 |
| [33] | OASIS | RF | 89.29 | - | 89 |
| | | XGB | 89.39 | - | 89 |
| | | GB | 91.02 | - | 91 |
| | | Voting 1 (Soft) | 91.17 | - | 91 |

Efforts are underway to detect AD early, aiming to slow the abnormal brain degeneration, lower healthcare costs, and enhance treatment outcomes. The utilization of ML techniques has demonstrated considerable promise in the categorization of dementia, a multifaceted neurological condition that impacts cognitive abilities. By utilizing sophisticated algorithms and computational methodologies, ML models possess the capacity to examine a wide range of data sources and assist in the timely identification, prediction, and tailored therapeutic strategies for individuals affected by dementia. Several researchers already deployed various ML models to classify dementia patient. Table 1 summarizes some ML models deployed in dementia diagnosis.

## 3.2 Heart Disease Detection

Machine learning (ML) approaches have been extensively used by researchers and practitioners to identify cardiac disease [33, 34]. For instance, a neurofuzzy-integrated system was developed in [33] for detecting heart disease that achieved accuracy of approximately 89%. Yet, the study's primary limitation is the absence of a well-defined clarification regarding the performance of their suggested technique across diverse scenarios like multiclass classification, extensive data analysis, and addressing imbalanced class distribution. Furthermore, there's a notable omission of dialogue regarding the model's trustworthiness and interpretability, a factor progressively vital in medical domains to enhance comprehensibility for non-medical individuals. In [35], researchers introduced a deep CNN to detect irregular cardiac sounds. They optimized the loss function to enhance sensitivity and specificity on the training dataset. This model underwent testing in the 2016 Physio Net computing competition, yielding a final prediction specificity of 95% and sensitivity of 73% [35].

Furthermore, deep learning (DL) algorithms have garnered attention in cardiac disease detection. In [36], a DL-based technique was developed for diagnosing cardiotocographic fetal health based on multiclass morphologic patterns. This model aimed to categorize patterns in individuals with pregnancy complications. Initial computational results displayed an accuracy of 88.02%, precision of 85.01%, and F-score of 85% [36]. Overfitting was addressed using various dropout strategies, leading to an increased training time, which they noted as a trade-off for achieving heightened accuracy. Liu et al. (2012) employed Support Vector Machine (SVM) to create predictive systems for cardiac arrest within 72 hours [37]. In a Cleveland dataset study, Shah et al. (2020) compared SVM, Random Forest (RF), Ordinal Regression, Logistic Regression (LR), and Naive Bayes (NB) for heart disease detection, with SVM yielding 95% accuracy [38]. Besides SVM and CNN, other algorithms like ensemble learning [39], k-Nearest Neighbors (kNN) [40], Decision Trees (DT) [41], Linear Discriminant Analysis (LDA) [42], and Bayesian Networks (BN) [43] were also employed in heart disease prediction. However, recent studies highlight Generative Adversarial Network (GAN) superiority for both balanced and imbalanced datasets. Researchers have introduced GAN-based models [44, 45, 46]. Wang et al. (2021) introduced CAB, a GAN-based approach addressing imbalance-related issues, achieving 99.71% accuracy in arrhythmia patients [44]. Rath et al. (2021) combined Long Short-Term Memory (LSTM) with GAN, accurately detecting heart disease patients from the MIT-BIH dataset with up to 99.4% accuracy [47].

These recent developments in GAN-based approaches showcase their potential in improving the accuracy and performance of machine learning models for cardiac disease diagnosis. The integration of GANs with other machine learning techniques holds promise for addressing imbalance-related challenges and achieving high accuracy in predicting heart diseases. Further research in this area is expected to enhance the capabilities of ML models to detect heart disease and contribute to more effective healthcare diagnostics.

Despite the wide adoption of ML applications in heart disease diagnosis, there is a lack of research addressing the challenges related to unbalanced data in multiclass classification. Additionally, most models lack sufficient explain ability during the final prediction, which hampers their understanding and trustworthiness. Further research is needed to address these issues and improve the transparency and robustness of ML-based cardiac disease detection systems.

## 3.3 Kidney Disease Detection

Chronic Kidney Disease (CKD) refers to a condition wherein the kidneys experience damage, leading to an impaired blood filtration process. The kidneys' primary function involves extracting excess water and waste from the blood to generate urine. In cases of CKD, the kidneys fail to effectively eliminate waste, resulting in its accumulation within the body. This ailment earns its "chronic" status due to the gradual and extended nature of the damage it inflicts over time. CKD stands as a prevalent global health concern, potentially giving rise to various health complications. The origins of CKD are diverse, encompassing factors like diabetes, elevated blood pressure, and heart disease.

Firstly, in [48], the authors conducted their research on clinical and blood biochemical measurements from 551 patients with proteinuria. Several ML models, including RF, extreme gradient boosting (XGB), LR, elastic net (ElasNet), lasso and ridge regression, k-NN, SVM, and artificial neural network (ANN), were compared for CKD risk prediction. The models ElasNet, lasso, ridge, and LR showed superior predictive performance, achieving a mean AUC and precision above 87% and 80%, respectively. LR ranked first, attaining an AUC of 87.3%, with a recall and specificity of 83% and 82%, respectively. ElasNet achieved the highest recall (0.85), while Extra Gradient Boosting (XGB) demonstrated the highest specificity (0.83). In a separate investigation [49], researchers employed SVM, AdaBoost, LDA, and gradient boosting (GBoost) algorithms to formulate accurate models for CKD prediction, utilizing a dataset from the UCI machine learning repository. The gradient boosting classifier attained the highest accuracy of 99.80%. In [50], authors concentrated on the CKD dataset, employing LR, Decision Tree (DT), and k-NN algorithms to develop three distinct CKD prediction models. LR exhibited the highest accuracy at 97%, outperforming DT (96.25%) and k-NN (71.25%). Similarly, another study [51] evaluated Naïve Bayes (NB), RF, and LR models for CKD risk prediction, achieving respective accuracies of 93.9%, 98.88%, and 94.76% on the dataset. Furthermore, in [52], a system for CKD risk prediction was proposed using data from 455 patients and real-time dataset. RF and ANN were trained and tested with 10-fold cross-validation, achieving accuracies of 97.12% and 94.5%, respectively. ANN and RF was deployed on CKD datasets having data of 455 instances with 25 features in [53]. The most significant features were collected using Chi-Square Test. The accuracy obtained by RF and ANN was 97.12% and 94.5%, respectively. A machine learning-based model was created in [54] with the aim of predicting chronic kidney disease (CKD) in patients. The model's performance was evaluated on two sets of data: one containing all attributes and another containing only selected features. Within the realm of feature selection methods, three common approaches are often employed: Wrapper, Filter, and Embedded. These methods serve the purpose of identifying and selecting the most crucial features for a given task or problem. The model was trained using various machine learning classifiers, including Artificial Neural Networks (ANN), C5.0, Logistic Regression (LR), Linear Support Vector Machine (LSVM), K-Nearest Neighbors (KNN), and Random Forest (RF). Based on the experimental findings, it was observed that the LSVM algorithm attained the maximum level of accuracy, reaching 98.86%, when applied to the SMOTE technique with all features included. SMOTE is widely regarded as an effective method for addressing class imbalance in datasets. The utilization of SMOTE in conjunction with feature selection by LASSO regression yielded superior outcomes compared to the LASSO regression model without the implementation of SMOTE [54].

In their study, Xiao et al. [55] utilized a dataset of 551 patients and implemented nine distinct machine learning algorithms. These algorithms included XGBoost, logistic regression, lasso regression, support vector machine, random forest, ridge regression, neural network, Elastic Net, and KNN. The researchers conducted an evaluation of many performance metrics, including accuracy, ROC curve, precision, and recall. The results indicated that the linear model exhibited the highest level of accuracy. Sujata Drall et al. [56] worked with the UCI-provided CKD dataset containing 400 instances and 25 attributes. First, data was pre-processed, then missing data was identified and replaced with zero, and the dataset was transformed and applied. After pre-processing, the authors employed an algorithm for significant attributes and identified the five most significant features, followed by the classification algorithms Nave Bayes and KNN. The obtained result KNN was the most accurate. Furthermore, a research study [57] employed classifiers such as extra-trees (ET), AdaBoost, KNN, GBoost, XGB, DT, Gaussian Naïve Bayes (NB), and RF. Among them, KNN and Extra Trees classifiers (ET) showed the best performance with accuracies of 99% and 98%, respectively. Highest Precision of 99.11% was achieved using ET and KNN.

In addition, ANN-based regression analysis for managing sparse medical datasets was proposed in [58]. To improve upon the pre-existing radial basis function (RBF) input-doubling technique, they incorporated new variables into the output signal calculation algorithm. Similarly, in [59], a new input doubling method based on the classical iterative RBF neural network was designed. The highest accuracy of the proposed method was validated by experimenting with a small medical dataset, using Mean Absolute Error and Root Mean Squared Error. In [60], an innovative method for data augmentation with enhanced disease categorization was implemented that was based on generative adversarial networks (GAN). Experiments were conducted on the NIH chest X-ray image dataset, and the test accuracy of CNN model was 60.3%. However, the online GAN-augmented CNN model showed improved performance, achieving a test accuracy of 65.3%. In [61], a methodology based on supervised learning was presented, focusing on developing efficient models for predicting the risk of chronic kidney disease (CKD) occurrence. The study mainly concentrated on probabilistic, tree-based, and ensemble learning-based models. Several algorithms were evaluated, including SVM, Logistic Regression (LR), Stochastic Gradient Descent (SGD), Artificial Neural Network (ANN), and k-NN.

### 3.4 Breast Cancer Detection

Breast cancer is the leading cancer in females worldwide, caused by abnormal growth of cells in the breast. Various techniques, including breast screening or mammography, have been introduced for accurate diagnosis. Mammography uses X-rays to check the nipple status of women, but early detection of small cancer cells remains challenging. Machine learning, deep learning, and bio-inspired computing techniques have been applied in medical prognoses, but none has consistently provided accurate results. Mammography requires doctors to analyze a large volume of imaging data, reducing accuracy and leading to time-consuming procedures with potential for misdiagnosis. As medical research advances, new systems are being developed for improved breast cancer detection. A denotes Accuracy, P denotes Precision, SP denotes Specificity and Se denotes Sensitivity. Table 2 summarizes the performance of some ML models in breast cancer classification.

**Table 2: State of art approaches for applying ML in Breast Cancer Classification**

| Ref. | Dataset | Models | A (%) | P (%) | SP (%) | SE (%) |
|---|---|---|---|---|---|---|
| [62] | WBC | SVM | - | - | 92.68% | 94.44% |
|  |  | LR | - | - | 90.48% | 94.37% |
|  |  | DT | - | - | 92.31% | 91.89% |
|  |  | RF | - | - | 94.59% | 90.79% |
|  |  | DNN | - | - | 91.11% | 98.53% |
| [63] | WBC | NB | 93% | 90% | - | 90% |
|  |  | LR | 97% | 100% | - | 92% |
| [64] | WBC | SVM | 97.14% | 95.65% | 92.3% | 100% |
|  |  | KNN | 97.14% | 97.82% | 95.83% | 97.82% |
|  |  | RF | 95.71% | 97.77% | 95.83% | 95.68% |
|  |  | LR | 95.71% | 97.82% | 95.65% | 95.74% |
| [65] | WDBC (569) | KNN | 96% | 93% (B), 100% (M) | - | 100%, 89% |
|  |  | SVM | 95% | 97%, 92% | - | 94%, 96% |
|  |  | DT | 97% | 97%, 98% | - | 99%, 96% |
|  |  | NB | 90% | 92%, 88% | - | 97%, 89% |
|  |  | LR | 96% | 97%, 96% | - | 97%, 96% |
| [66] | WBC (699) | MLP | 95.44% | 95.4% | - | 95.4% |
|  |  | Voted Perceptron | 90.98% | 89.9% | - | 88.2% |
| [67] | WBC (699) | KNN | 97.51% | - | - | - |
|  |  | NB | 96.19% | - | - | - |

## 4. CHALLENGES & FUTURE DIRECTIONS

ML-based applications have been widely employed in illness detection; nevertheless, the implementation of these applications in healthcare as practical tools presents several problems for researchers and practitioners. Even though numerous hospitals and healthcare institutions have collected extensive patient data, the availability of real-world data for worldwide research purposes is limited due to the constraints imposed by data privacy regulations. Often, clinical data is subject to noise or missing values, resulting in a significant time investment required to render such data trainable. The problem of adversarial attack is a significant challenge within the context of illness datasets. The utilization of machine learning models in the development of illness diagnosis models carries the potential for significant harm in the event of misclassification pertaining to a specific disease. For example, the misdiagnosis of a patient with stomach cancer as a non-cancer patient can have significant consequences. One of the primary issues associated with the machine learning (ML) model pertains to its tendency to frequently misidentify a region as diseased, hence leading to erroneous outcomes. The majority of machine learning models, including logistic regression (LR), exhibited high levels of performance when trained on labelled data. Nevertheless, the performance of comparable algorithms experienced a notable decrease when exposed to the unlabeled data. However, it should be noted that certain widely-used algorithms, such as K-means clustering, SVM, and K-Nearest Neighbors (KNN), may experience a decline in performance when applied to multidimensional data.

The issues discussed in the preceding part may provide valuable insights for future scholars and practitioners, guiding their future endeavors. The utilization of generative adversarial networks (GANs) has gained significant prominence within the realm of deep learning. By employing this methodology, it becomes feasible to produce artificial data that has a striking resemblance to authentic data. Hence, the utilization of GANs could potentially serve as a viable solution for addressing challenges related to limited availability of data. The progression of contemporary technology has facilitated the acquisition of data with high resolutions and multiple attributes. Although the conventional ML approach may not yield satisfactory results when applied to high-quality data, employing a mix of many ML models could prove to be a viable solution for effectively managing such data with many dimensions.

## 5. CONCLUSION

Machine learning has the potential to bring about numerous technological revolutions in the healthcare industry. It has the potential to enhance diagnostic accuracy, facilitate the discovery of trends and patterns in patient data, streamline administrative processes, and make possible patient-specific treatment plans. Both supervised and unsupervised learning have their advantages and disadvantages in the medical field. The task at hand, the amount of data at hand, and the resources at your disposal will all dictate the style of learning you employ. Machine learning will become increasingly important in healthcare as data volumes increase. Further investigation into the constraints discussed in the paper's final two sections would be very welcome. Future MLBDD research could also center on issues such optimizing big data sets that include numerical, categorical, and picture data, as well as multiclass classification with highly imbalanced data and highly missing data, and the explanation and interpretation of multiclass data classification utilizing XAI.